# BRAIN MRI SUPER RESOLUTION USING 3D DEEP DENSELY CONNECTED NEURAL NETWORKS

*Yuhua Chen[1,2], Yibin Xie[2], Zhengwei Zhou[1,2], Feng Shi[2], Anthony G. Christodoulou[2], Debiao Li[1,2]*

[1]Department of Bioengineering, UCLA, Los Angeles, California, USA
[2]Biomedical Image Research Institute, Cedars-Sinai Medical Center, Los Angeles, California, USA

## ABSTRACT

Magnetic resonance image (MRI) in high spatial resolution provides detailed anatomical information and is often necessary for accurate quantitative analysis. However, high spatial resolution typically comes at the expense of longer scan time, less spatial coverage, and lower signal to noise ratio (SNR). Single Image Super-Resolution (SISR), a technique aimed to restore high-resolution (HR) details from one single low-resolution (LR) input image, has been improved dramatically by recent breakthroughs in deep learning. In this paper, we introduce a new neural network architecture, 3D Densely Connected Super-Resolution Networks (DCSRN) to restore HR features of structural brain MR images. Through experiments on a dataset with 1,113 subjects, we demonstrate that our network outperforms bicubic interpolation as well as other deep learning methods in restoring 4x resolution-reduced images.

## 1. INTRODUCTION

Medical images in high spatial resolution (HR) produce abundant structural details, enabling accurate image analysis and quantitative measurement. However, high spatial resolution in MRI typically comes at the expense of longer scan time, less spatial coverage, and lower signal to noise ratio (SNR) [1]. If we could reconstruct a high-resolution (HR) image from a low-resolution (LR) input, we can potentially achieve larger spatial coverage, higher SNR and better spatial resolution in a shorter scan. A simplistic approach is to interpolate LR images into HR. However, interpolation methods fail to recover the loss of high-frequency information like fine edges of objects. Another approach is to scan multiple LR images and combine them into a single HR image. Unfortunately, this is not robust to interscan motion, and is neither time- nor cost-optimal in practice. Therefore, a SISR [2] technique which needs only one LR scan to provide an HR output is an attractive approach to address this problem.

Before data-driven machine learning was widely used, most applications [3] of SISR took the form of an optimization problem to minimize the cost between observed LR image and the model estimation, typically with some form of regularization terms. However, those non-learning methods generally have a limitation that they require sound prior knowledge about the data representation. For instance, the use of a total variation regularization term implicitly assumes that the image should be piecewise constant, which does not always hold for images with abundant structural details.

In contrast, a learning-based method does not require the assumption of the data distribution but learns the prior information directly from a set of examples [4]. Recent state-of-the-art methods with deep learning techniques have shown a great performance improvement on natural images. Among those approaches, Super-Resolution Convolutional Neural Networks (SRCNN) [5] and its faster version FSRCNN [6] have received substantial attention because of their simple network structure and high restoration accuracy.

However, these previous deep-learning approaches still have limitations. First, many medical images are 3D volumes, but 2D super-resolution networks like the original FSRCNN work slice-by-slice without taking advantage of continuous structures in 3D. A 3D model would be preferable, as it directly extracts 3D image features, considering objects across multiple slices. Second, FSRCNN stacks multiple convolutional neural networks. Direct conversion into 3D may result in a large number of parameters and thus faces challenges in memory allocation. Finally, its structure could also be further improved in terms of efficiency.

In this paper, we propose 3D Densely Connected Super-Resolution Networks (DCSRN), derived from a recent development in neural networks, Densely Connected Convolutional Networks [7]. Experiments are performed on a large brain MRI dataset from the human connectome project (HCP) [8]. We show that in brain MRI SR, 3D neural networks outperformed their 2D counterparts, and that our 3D-DCSRN also outperformed a previous method for 3D FSRCNN. In addition, in our study, the low-resolution image is created by truncating k-space instead of downsampling in the image domain, which more realistically represents the acquisition of low-resolution MRI scans.

## 2. SUPER-RESOLUTION NEURAL NETWORKS

### 2.1. Background

As SISR is an estimation process in image space to turn LR images $Y$ into ground truth HR images $X$, the relation between $X$ and $Y$ can be interpreted as:

$$Y = f(X) \qquad (1)$$

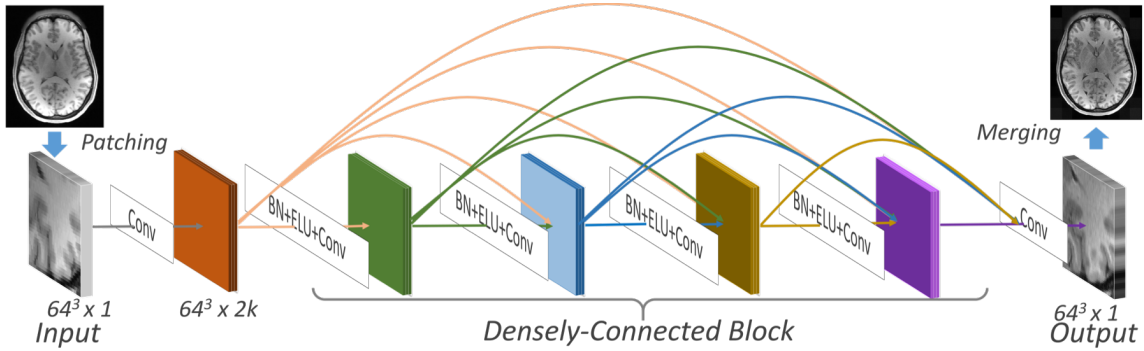

**Figure 1**. Framework of the proposed 3D Densely Connected Super-Resolution Networks (DCSRN).

where the $f$ is an arbitrary continuous transformation function that downgrades the image $X$. The SR process will be to find some $g(\cdot) \approx f^{-1}(\cdot)$, where $f^{-1}$ is the inverse mapping function of $f$, to reconstruct an HR image as:

$$X = g(Y) = f^{-1}(Y) + r \qquad (2)$$

where $r$ is the reconstruction residual.

A learning-based SR typically has three steps to restore $X$: 1) extract image features from $Y$, 2) map the feature vector to a feature space, and 3) reconstruct $X$ from the feature space.

It has been shown that convolutional neural networks [9] can handle those operations naturally [5]. By minimizing the difference between reconstructed images and ground truth images during the training process, the model learns the transformation from LR to HR by those three steps.

### 2.2. Proposed 3D Densely Connected Super-Resolution Networks (DCSRN)

Though FSRCNN has a significant improvement in speed over SRCNN, recent studies [7, 10] showed that more sophisticated network structures with skip connections and layer reusing benefit not only performance and speed, but also reduces training time. Inspired by the Densely Connected Network (DenseNet) used in object recognition [7], we propose a new SR network, referred to as DCSRN. There are three major benefits to use DCSRN: 1) faster training—each path in the proposed network is much shorter, so back-propagation is more efficient; 2) a light-weight model—thanks to weight sharing, the model is small and efficient; 3) less overfitting during training—the number of parameters is greatly reduced and features are reused heavily, so it is difficult for overfitting to occur. The network structure of DCSRN is shown in Fig. 1. Patches are extracted from the whole 3D image and fed into the network. A convolutional layer with kernel size of 3 and filter number of 2k is applied to the input image before a densely-connected block with 4 units, each which has a batch normalization layer and an exponential linear units (ELUs) activation followed by a conv layer with $k$ filters. A conv layer is used to provide final SR output.

### 2.3. 3D model vs 2D model

In many studies, a 2D model was directly applied to 3D medical images slice-by-slice or combining results from coronal, axial, and sagittal views. However, as medical images carry structural information in 3D form, a 3D model is a natural way to learn richer knowledge. For example, a small blood vessel may cast its edge to a neighboring slice; when a 2D SR model processes the adjacent slice, it might be difficult to determine whether this small fluctuation is part of the vessel or just the noise, potentially resulting in noise enhancement in the output.

## 3. EXPERIMENTS

### 3.1. Evaluation Setting

For evaluation, we create LR images from the HR images, and the SR results from LR images are then compared with the HR ground truth to evaluate the performance.

Unlike previous SR approaches [2,11] where the LR images were created in image domain through a Gaussian blurring followed by shrinking image size in all dimensions, we obtained LR images by: 1) applying the FFT to HR images, converting the original image into k-space data; 2) degrading the resolution by zeroing the outer part of the 3D k-space along two axes representing two MR phase encoding directions; 3) applying the inverse FFT. This process results in LR images at the same image size as the HR images, avoiding checkerboard artifacts [12]. This mimics the real image acquisition process where a low-resolution MRI is scanned by reducing acquisition lines in phase and slice encoding directions. The missing data is in k-space, thus the blurring pattern is different from simply reducing image size in the image domain. Our process more faithfully follows the real LR MRI acquisition process.

### 3.2. Dataset

In a previous experiment with a small dataset of 21 images [11], 3D SRCNN demonstrated superior performance against the 2D version. In order to better demonstrate the network's generalization, we chose a large publicly accessible brain MRI database acquired from multiple centers, the human connectome project [8], and employ 1113 subjects' brain T1w structural images. These 3D MPRAGE images were obtained from Siemens 3T platforms using a 32-channel head coil. The matrix size is 320x320x256. The spatial resolution is 0.7 mm isotropic. The whole dataset was randomly split into 7:1:1:1 ratio as 780 training, 111 validation for optimizing network weights, 111 evaluation for choosing hyperparameters, and 111 test samples for unbiasedly performance checking. The original images were treated as the ground-truth HR images, and then degraded to LR ones, lowering the spatial resolution by a factor of 2 in each phase encoding direction (for a total factor of 4).

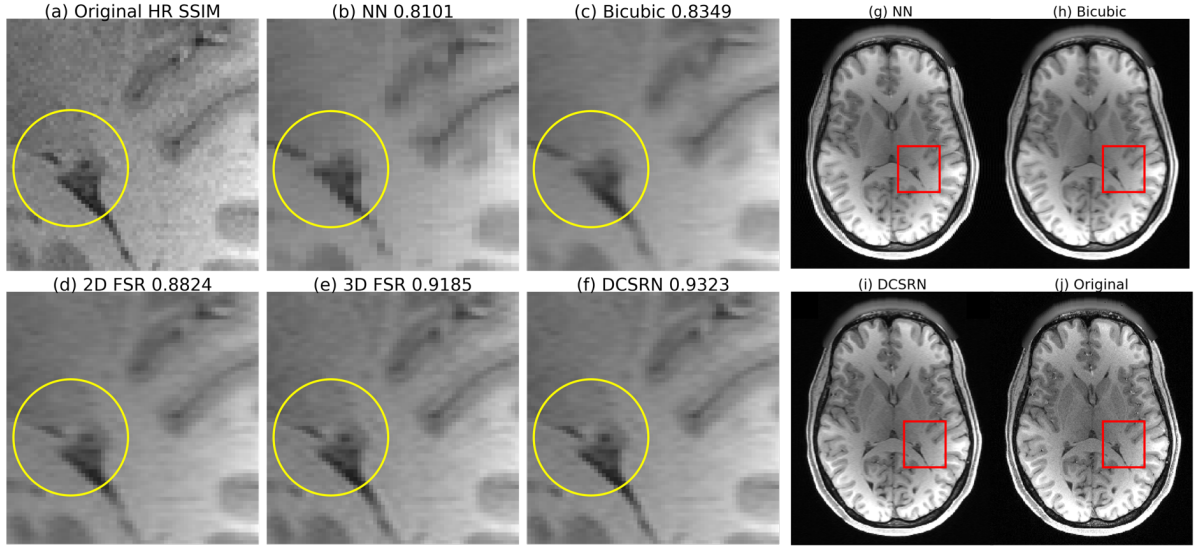

**Figure 2**. One randomly selected sample, zoom-ins of the red region are shown alongside and SSIMs of this image are on the top. A small vessel in the yellow circle is blurred out in (b) and (c), partially recovered in (d) and (e), and DCSRN (f) preserves more details.

### 3.3. Patching, Merging and Data Augmentation

Each subject's 3D image was split into 64x64x64 cubes. The locations of the cubes were randomly selected in each training step within the whole 3D image volume, acting as random translation for data augmentation. To compare against 2D networks, each 3D cube was again split into 64 2D patches. A batch of two 3D cubes were fed into the network during training. In the testing phase, the whole SR volumes were merged by averaging the model's SR output cubes in a 3D sliding window manner. In each sliding step, the window shifted by half of the cube size.

### 3.4. Training

We implemented all the models in Tensorflow [13]. For the proposed DCSRN, the densely-connected block has four 3x3x3 convolutional layers with 48 filters as the first layer output in a growth rate (k) of 24, which gave us the best results in this work. For the comparison methods, we selected the hyperparameter according to [6] for 2D FSRCNN (Table 1). We kept the same layer settings but extended all 2D convolution to 3D for the 3D-FSRCNN.

Adam optimizer with a learning rate $10^{-5}$ was used to minimize the L2 loss (mean squared error) between the network output SR images and the corresponding HR ground truth during training. All models were trained from scratch on a workstation with a Nvidia GTX 1080 TI GPU for approximately 72 hours, after which little improvement was observed.

Our program saved the model checkpoint with best validation loss during training, and applied that model to the test set for performance analysis.

| | Nearest Neighbor | | | Bicubic Interpolation | | | 2D FSRCNN | | | 3D FSRCNN | | | 3D DCSRN | | |
|---|---|---|---|---|---|---|---|---|---|---|---|---|---|---|---|
| | SSIM | PSNR | NRMSE | SSIM | PSNR | NRMSE | SSIM | PSNR | NRMSE | SSIM | PSNR | NRMSE | SSIM | PSNR | NRMSE |
| mean | 0.8131 | 28.39 | 0.2049 | 0.8382 | 29.21 | 0.1868 | 0.8836 | 31.28 | 0.1467 | 0.9166 | 33.86 | 0.1093 | **0.9312** | **35.05** | **0.0954** |
| std | 0.0086 | 0.85 | 0.0086 | 0.0079 | 0.86 | 0.0082 | 0.0075 | 0.78 | 0.0048 | 0.0066 | 0.79 | 0.0041 | 0.0064 | 0.84 | 0.0041 |
| min | 0.7873 | 26.54 | 0.1852 | 0.8145 | 27.36 | 0.1674 | 0.8635 | 29.75 | 0.1340 | 0.8997 | 32.05 | 0.0998 | 0.9156 | 32.97 | 0.0885 |
| median | 0.8134 | 28.30 | 0.2033 | 0.8386 | 29.10 | 0.1854 | 0.8842 | 31.35 | 0.1471 | 0.9175 | 33.88 | 0.1085 | **0.9320** | **35.05** | **0.0944** |
| max | 0.8318 | 30.54 | 0.2263 | 0.8542 | 31.32 | 0.2083 | 0.9005 | 33.57 | 0.1593 | 0.9315 | 36.18 | 0.1227 | 0.9468 | 37.57 | 0.1090 |

**Table 1**. The results of SSIM, PSNR and normalized root mean squared error for a downgrade factor of 2x2 between nearest-neighbor, bicubic interpolation, 2D FSRCNN, 3D FSRCNN and 3D DCSRN. 3D DCSRN has highest average similarity scores in SSIM and PSNR and lowest mean voxel-wise intensity difference to ground truth HR images.

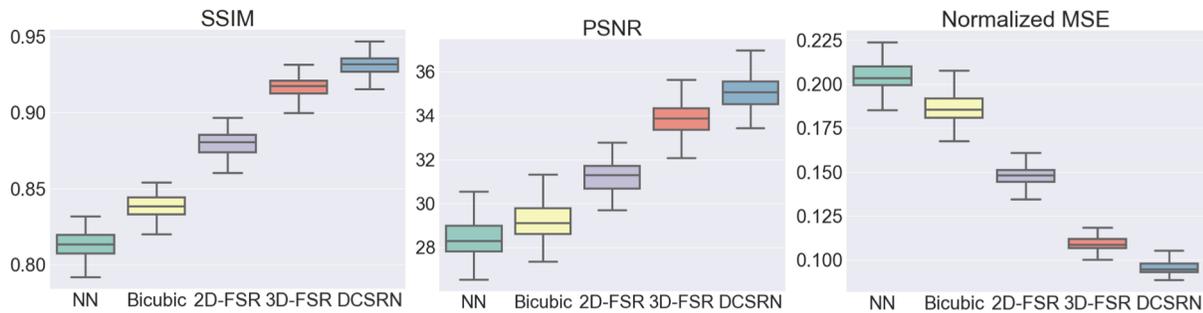

**Figure 3**. Boxplot of results of SSIM, PSNR and NRMSE between different methods. DCSRN scores the best in all three metrics.

## 4. RESULTS

The results from different methods of a sample case are shown in Fig 2. To quantitatively measure the results, we computed three image metrics: Structural Similarity Index (SSIM) [14], peak signal to noise ratio (PSNR) and normalized root mean squared error (NRMSE), between the SR images and the HR ground truth.

The comparison methods were nearest neighbor up-sampling, bicubic interpolation, and three deep learning models: 2D-FSRCNN, 3D-FSRCNN, and DCSRN. Table 1 provides a summary of quantitative analysis. Figure 3 is the boxplot of the results. In all 3 metrics, all deep learning models show better performance than simple interpolation methods by a large margin. Additionally, 3D FSRCNN outperforms 2D FSRCNN, and our 3D DCSRN has the best performance over all ($p<0.01$ from two-sample t-tests).

In terms of the speed, we observed that DCSRN trained 4x faster than 3D FSRCNN. In testing, to process a patient's 3D image set with patch size 64x64x64, DCSRN took 23.31s, 2D FSRCNN took 35.52s, and 3D FSRCNN took 63.95s. The proposed DCSRN was the fastest among them.

## 5. CONCLUSION

In this paper, we demonstrated a novel convolutional neural network DCSRN for SISR of 3D brain MRI. Although it is possible that hyperparameters of the model may affect the performance, with the limited time and resource, we could not explore the combination in exhaustively. A study on hyperparameters searching on various structures will be one potential future work. However, the results showed that compared with popular interpolation and previous deep learning methods, the new model produced significantly better quality SR images and did so more efficiently.

**Acknowledgement** Data were provided in part by the Human Connectome Project, WU-Minn Consortium (Principal Investigators: David Van Essen and Kamil Ugurbil; 1U54MH091657) funded by the 16 NIH Institutes and Centers that support the NIH Blueprint for Neuroscience Research; and by the McDonnell Center for Systems Neuroscience at Washington University.